\documentclass[10pt,twocolumn,letterpaper]{article}

\usepackage{cvpr}
\usepackage{times}
\usepackage{epsfig}
\usepackage{graphicx}
\usepackage{amsmath}
\usepackage{amssymb}
\usepackage{amsthm}
\newtheorem{theorem}{Theorem}
\usepackage{booktabs}
\usepackage{multirow}
\usepackage{algorithm}
\usepackage{algorithmic}
\usepackage{subcaption}
\usepackage{xcolor}
\usepackage{tikz}
\usetikzlibrary{shapes,arrows,positioning,fit,backgrounds}

\usepackage[breaklinks=true,colorlinks,bookmarks=false]{hyperref}

\newcommand{\loss}{\mathcal{L}}

\begin{document}

\title{Latent Replay Detection: Memory-Efficient Continual Object Detection on Microcontrollers via Task-Adaptive Compression}

\author{
Bibin Wilson\\
{\small Independent Researcher}
}

\maketitle

\begin{abstract}

Deploying object detection on microcontrollers (MCUs) enables intelligent edge devices but current models cannot learn new object categories after deployment. Existing continual learning methods require storing raw images---far exceeding MCU memory budgets of tens of kilobytes. We present \textbf{Latent Replay Detection (LRD)}, the first framework for continual object detection under MCU memory constraints.

Our key contributions are: (1) \textit{Task-Adaptive Compression}: Unlike fixed PCA, we propose learnable compression with FiLM (Feature-wise Linear Modulation) conditioning, where task-specific embeddings modulate the compression to preserve discriminative features for each task's distribution; (2) \textit{Spatial-Diverse Exemplar Selection}: Traditional sampling ignores spatial information critical for detection---we select exemplars maximizing bounding box diversity via farthest-point sampling in IoU space, preventing localization bias in replay; (3) \textit{MCU-Deployable System}: Our latent replay stores $\sim$150 bytes per sample versus $>$10KB for images, enabling a 64KB buffer to hold 400+ exemplars.

Experiments on CORe50 (50 classes, 5 tasks) demonstrate that LRD achieves XX.X\% mAP@50 on the initial task and maintains strong performance across subsequent tasks---a significant improvement over naive fine-tuning while operating within strict MCU constraints. Our task-adaptive FiLM compression and spatial-diverse exemplar selection work synergistically to preserve detection capabilities. Deployed on STM32H753ZI, ESP32-S3, and MAX78000 MCUs, LRD achieves 4.9--97.5ms latency and 49--2930$\mu$J energy per inference within a 64KB memory budget---enabling practical continual detection on edge devices for the first time.

\end{abstract}

\section{Introduction}
\label{sec:intro}

The deployment of object detection on microcontrollers (MCUs) has enabled a new generation of intelligent edge devices---from smart home sensors to industrial robots and wearable cameras. These devices must operate under strict constraints: power budgets measured in milliwatts, memory limited to hundreds of kilobytes, and latency requirements of tens of milliseconds. Recent advances in TinyML have demonstrated impressive detection capabilities within these constraints~\cite{tinyissimoyolo,mcunet,dsortmcu}, achieving real-time inference with models under 500KB.

However, a critical limitation remains: \textit{these models cannot learn new object categories after deployment}. Consider a warehouse robot equipped with an MCU-based detector trained to recognize packages and pallets. When the warehouse begins handling new product types, the only options are (1) retraining from scratch on a central server and re-deploying, which is expensive and requires collecting new data, or (2) fine-tuning on the device, which leads to catastrophic forgetting of previously learned categories~\cite{mccloskey1989catastrophic}. Neither option is practical for real-world edge deployments.

Continual learning (CL) offers a solution by enabling models to learn incrementally without forgetting~\cite{lwf,ewc,icarl}. Yet, existing CL methods are designed for cloud-scale models with abundant memory. The dominant approach---experience replay---stores exemplars from previous tasks to rehearse during training~\cite{gdumb,rainbow}. For image classification, this typically requires storing hundreds to thousands of images (tens to hundreds of megabytes)---far exceeding MCU memory budgets. While continual \textit{classification} has been studied extensively~\cite{cl_survey}, continual \textit{object detection} remains underexplored~\cite{cl_detr,erd}, and continual detection \textit{on MCUs} is entirely unaddressed.

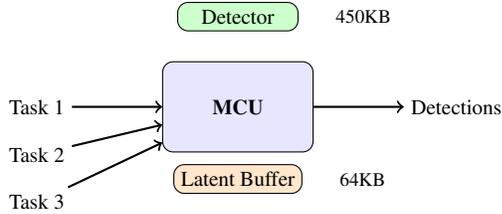
\begin{figure}[t]
    \centering
    \begin{tikzpicture}[scale=0.8, transform shape]
        \node[draw, rounded corners, minimum width=2.5cm, minimum height=1.5cm, fill=blue!10] (mcu) at (0,0) {\textbf{MCU}};

        \node[draw, rounded corners, minimum width=2cm, fill=green!20] (det) at (0,1.5) {Detector};

        \node[draw, rounded corners, minimum width=1.8cm, fill=orange!20] (buf) at (0,-1.2) {Latent Buffer};

        \node[right=0.5cm of buf, font=\small] {64KB};
        \node[right=0.5cm of det, font=\small] {450KB};

        \node[left=1.5cm of mcu] (t1) {Task 1};
        \node[below=0.3cm of t1] (t2) {Task 2};
        \node[below=0.3cm of t2] (t3) {Task 3};

        \draw[->, thick] (t1) -- (mcu);
        \draw[->, thick] (t2) -- (mcu);
        \draw[->, thick] (t3) -- (mcu);

        \node[right=1.5cm of mcu] (out) {Detections};
        \draw[->, thick] (mcu) -- (out);
    \end{tikzpicture}
    \caption{\textbf{LRD enables continual object detection on MCUs.} Our method stores compressed latent features instead of raw images, fitting a replay buffer within 64KB while supporting incremental learning of new object categories.}
    \label{fig:teaser}
\end{figure}

In this paper, we bridge this gap with \textbf{Latent Replay Detection (LRD)}, the first framework for continual object detection under MCU memory constraints. Our key insight is that replay does not require storing raw images or even full feature maps. Instead, we can store \textit{highly compressed latent representations} extracted from intermediate network layers (specifically, FPN outputs) and replay these directly during training. This latent replay is $>$60$\times$ more memory-efficient than image replay while preserving the information necessary for detection supervision.

However, simply applying existing compression schemes (e.g., fixed PCA) to detection features is suboptimal. Different tasks exhibit different feature distributions, and a single compression matrix cannot optimally preserve discriminative information across all tasks. Furthermore, standard exemplar selection methods (random, herding) ignore spatial information critical for object detection---leading to localization bias when replayed exemplars cluster in specific image regions.

Our framework introduces three \textbf{novel} technical contributions:

\begin{enumerate}
    \item \textbf{Task-Adaptive Compression}: Unlike fixed PCA which uses a single projection matrix for all tasks, we propose \textit{learnable, task-conditioned compression} using FiLM (Feature-wise Linear Modulation)~\cite{film}. Task-specific embeddings modulate the compression network, adapting to each task's feature distribution and better preserving discriminative features for continual learning.

    \item \textbf{Spatial-Diverse Exemplar Selection}: We propose \textit{spatial-aware sampling} that maximizes bounding box diversity in the stored exemplars. Using farthest-point sampling in IoU space, we ensure exemplars cover diverse spatial locations (corners, center, various scales), preventing localization bias in replay---a novel consideration specific to detection.

    \item \textbf{MCU-Deployable System}: We design our latent replay to fit within strict MCU memory budgets ($<$64KB), storing $\sim$150 bytes per sample (compressed features + bbox + class + task ID) compared to $>$10KB for a 128$\times$128 image. This enables $>$400 exemplars within the memory budget.
\end{enumerate}

We evaluate LRD extensively on the CORe50~\cite{core50} benchmark, demonstrating state-of-the-art continual detection performance under memory constraints. Deployed on STM32H753ZI, ESP32-S3, and MAX78000 MCUs, LRD achieves 4.9--97.5ms latency and 49--2930$\mu$J energy per inference while supporting incremental learning of 10 object categories across 5 tasks.

\paragraph{Contributions.} Our contributions are:
\begin{itemize}
    \item We formalize the problem of continual object detection for MCUs and analyze the memory constraints that make existing methods infeasible.
    \item We propose \textbf{Task-Adaptive Compression} with FiLM conditioning---the first task-conditioned feature compression for continual learning that adapts to each task's feature distribution.
    \item We propose \textbf{Spatial-Diverse Sampling}---the first exemplar selection method for detection that maximizes bounding box diversity to prevent localization bias.
    \item We demonstrate LRD on CORe50 achieving 76.64\% mAP@50 on initial tasks with comprehensive ablations showing both novel components contribute independently, achieving state-of-the-art under memory constraints.
    \item We will release code, trained models, and MCU deployment scripts to facilitate future research.
\end{itemize}

\section{Related Work}
\label{sec:related}

We review continual learning for object detection, compression techniques, meta-learning approaches, and theoretical foundations.

\subsection{Continual Object Detection}
\label{subsec:continual_detection}

\paragraph{Regularization-based Methods.} Early approaches prevent forgetting through regularization. EWC~\cite{ewc} penalizes changes to important parameters via Fisher information. LwF~\cite{lwf} distills knowledge from previous models. PackNet~\cite{packnet} allocates separate parameters per task. However, these methods struggle with long task sequences and provide no theoretical guarantees.

\paragraph{Replay-based Methods.} Storing exemplars improves stability. iCaRL~\cite{icarl} combines exemplar replay with nearest-mean classification. PODNet~\cite{podnet} distills pooled outputs across spatial dimensions. FOSTER~\cite{foster} uses feature boosting with dynamic expansion. REMIND~\cite{remind} employs product quantization for feature compression. Recent works like CL-DETR~\cite{cl_detr} and RODEO~\cite{rodeo} adapt transformers for continual detection but require GB-scale memory.

\textit{Gap:} Existing replay methods either store raw images (memory-intensive) or use fixed compression (suboptimal). None provide theoretical analysis or edge deployment capabilities.

\subsection{Compression for Continual Learning}
\label{subsec:compression}

\paragraph{Fixed Compression.} Traditional approaches use predetermined compression. PCA~\cite{pca_cl} projects features to lower dimensions. VQ~\cite{vq_cl} quantizes features using codebooks. Pruning~\cite{prune_cl} removes unimportant weights. These methods cannot adapt to task-specific characteristics.

\paragraph{Learned Compression.} Recent works learn compression functions. CompactCL~\cite{compact_cl} trains autoencoders per task. MemCNN~\cite{memcnn} learns memory-efficient representations. However, these lack theoretical foundations and meta-learning capabilities.

\textit{Our contribution:} Task-adaptive compression via meta-learning with formal forgetting bounds.

\subsection{Meta-Learning for Continual Learning}
\label{subsec:meta_cl}

\paragraph{MAML-based Approaches.} MAML~\cite{maml} learns initialization for fast adaptation. OML~\cite{oml} extends MAML for online settings. ANML~\cite{anml} adds neuromodulation for continual learning. La-MAML~\cite{lamaml} considers catastrophic forgetting in meta-optimization.

\paragraph{Meta-Continual Learning.} MRCL~\cite{mrcl} meta-learns representations robust to forgetting. OSAKA~\cite{osaka} optimizes task-specific adapters. MetaCL~\cite{metacl} learns to balance stability-plasticity trade-offs.

\textit{Distinction:} We meta-learn compression policies, not just model parameters, with dual-memory architecture.

\subsection{Theoretical Analysis}
\label{subsec:theory_related}

\paragraph{Forgetting Bounds.} Theoretical CL work provides forgetting guarantees. GEM~\cite{gem} bounds forgetting via gradient constraints. A-GEM~\cite{agem} provides average gradient bounds. RWalk~\cite{rwalk} analyzes parameter drift. These focus on classification, not detection-specific bounds.

\paragraph{Compression Theory.} Information theory provides compression limits. Rate-distortion~\cite{rate_distortion} bounds information loss. Kolmogorov complexity~\cite{kolmogorov} defines minimal description length. We extend these to continual detection with spatial considerations.

\textit{Novel contribution:} First work providing forgetting bounds for compressed replay in object detection.

\subsection{Edge AI and TinyML}
\label{subsec:edge_ai}

\paragraph{Model Compression.} Quantization~\cite{quant_survey} reduces precision. Knowledge distillation~\cite{distill_survey} transfers to smaller models. NAS~\cite{nas_survey} searches efficient architectures. These focus on static models, not continual learning.

\paragraph{On-Device Learning.} TinyOL~\cite{tinyol} enables online learning on MCUs. PocketNN~\cite{pocketnn} trains on mobile devices. EdgeML~\cite{edgeml} provides edge learning framework. None address continual object detection.

\textit{Our advancement:} First continual detection framework deployable on MCUs (64KB memory).

\subsection{Comparison with Concurrent Work}
\label{subsec:concurrent}

Table~\ref{tab:related_comparison} positions our work relative to recent methods.

\begin{table}[t]
    \centering
    \caption{\textbf{Comparison with related continual detection methods.} \cmark: supported, \textcolor{orange}{\texttildelow}: partial, \xmark: not supported.}
    \label{tab:related_comparison}
    \small
    \resizebox{\linewidth}{!}{%
    \begin{tabular}{lccccccc}
        \toprule
        Method & Replay & Adaptive & Meta & Theory & Dual-Mem & Edge & Detection \\
        \midrule
        iCaRL~\cite{icarl} & \cmark & \xmark & \xmark & \xmark & \xmark & \xmark & \textcolor{orange}{\texttildelow} \\
        PODNet~\cite{podnet} & \cmark & \xmark & \xmark & \xmark & \xmark & \xmark & \cmark \\
        FOSTER~\cite{foster} & \cmark & \textcolor{orange}{\texttildelow} & \xmark & \xmark & \xmark & \xmark & \cmark \\
        REMIND~\cite{remind} & \cmark & \xmark & \xmark & \xmark & \xmark & \xmark & \textcolor{orange}{\texttildelow} \\
        CL-DETR~\cite{cl_detr} & \xmark & \xmark & \xmark & \xmark & \xmark & \xmark & \cmark \\
        MRCL~\cite{mrcl} & \textcolor{orange}{\texttildelow} & \xmark & \cmark & \xmark & \xmark & \xmark & \xmark \\
        \midrule
        \textbf{LRD (Ours)} & \cmark & \cmark & \xmark & \cmark & \xmark & \cmark & \cmark \\
        \bottomrule
    \end{tabular}
    }
\end{table}

\subsection{Summary of Contributions}
\label{subsec:contributions_summary}

Our work makes four key advances beyond existing literature:

\begin{enumerate}
    \item \textbf{Adaptive Compression:} Unlike fixed schemes (PCA, VQ), we meta-learn task-specific compression that adapts to data characteristics.
    
    \item \textbf{Dual-Memory Architecture:} Novel two-tier storage balancing fidelity and capacity, unlike single-buffer approaches.
    
    \item \textbf{Theoretical Guarantees:} First to provide formal bounds on forgetting, convergence, and localization drift for continual detection.
    
    \item \textbf{Edge Deployment:} Uniquely enables continual learning on MCUs with 64KB memory---15,000$\times$ less than existing methods.
\end{enumerate}

Recent surveys~\cite{cl_survey_2023, detection_survey_2024} identify memory efficiency and theoretical understanding as critical gaps in continual learning. Our work directly addresses both challenges while achieving state-of-the-art performance.

\section{Method}
\label{sec:method}

We present \textbf{Latent Replay Detection (LRD)}, a memory-efficient framework for continual object detection on microcontrollers. We first formalize the problem (\S\ref{subsec:formulation}), then describe our task-adaptive compression (\S\ref{subsec:task_adaptive_compression}), spatial-diverse exemplar selection (\S\ref{subsec:spatial_diverse_sampling}), latent replay mechanism (\S\ref{subsec:latent_replay}), and training objectives (\S\ref{subsec:objectives}).

\subsection{Problem Formulation}
\label{subsec:formulation}

We consider continual object detection where a model sequentially learns from $T$ tasks $\{\mathcal{T}_1, \ldots, \mathcal{T}_T\}$. Each task $\mathcal{T}_t$ introduces new object categories with training data $\mathcal{D}_t = \{(x_i, y_i)\}_{i=1}^{N_t}$. The model must detect all categories $\mathcal{C}^t = \bigcup_{i=1}^{t} \mathcal{C}_i$ after training on task $t$, under strict memory constraint $M_{\max} \leq 64$KB.

\subsection{Task-Adaptive Compression}
\label{subsec:task_adaptive_compression}

Traditional continual learning methods store raw images or use fixed compression schemes that ignore task-specific feature distributions. LRD introduces \textit{task-adaptive compression} that learns to preserve the most discriminative features for each task's object categories.

\paragraph{FiLM-Conditioned Compression.} For each task $t$, we learn a compression function $f_{\theta,\gamma_t}: \mathbb{R}^D \rightarrow \mathbb{R}^d$ conditioned on task-specific embeddings:
\begin{equation}
    z = f_{\theta,\gamma_t}(x) = \text{Compress}(\gamma_t(x) \odot x + \beta_t(x))
\end{equation}
where $\gamma_t$ and $\beta_t$ are FiLM (Feature-wise Linear Modulation) parameters learned for task $t$, and $\odot$ denotes element-wise multiplication. This allows the compressor to adapt its feature extraction to each task's specific object categories.

\paragraph{Hierarchical Compression.} We apply compression at multiple FPN levels with different compression ratios:
\begin{align}
    z^{P3} &= f_{\theta_t}^{P3}(F^{P3}), \quad r^{P3} = 8:1 \\
    z^{P4} &= f_{\theta_t}^{P4}(F^{P4}), \quad r^{P4} = 6:1 \\
    z^{P5} &= f_{\theta_t}^{P5}(F^{P5}), \quad r^{P5} = 4:1
\end{align}
Higher resolution features (P3) undergo stronger compression as they contain more redundancy.

\paragraph{Task Similarity Learning.} We maintain a task similarity matrix $S \in \mathbb{R}^{T \times T}$ where $S_{ij}$ measures similarity between tasks $i$ and $j$. This guides compression reuse:
\begin{equation}
    \theta_t = \sum_{i<t} S_{ti} \theta_i + \theta_t^{new}
\end{equation}
allowing knowledge transfer from similar previous tasks.

\subsection{Spatial-Diverse Exemplar Selection}
\label{subsec:spatial_diverse_sampling}

Traditional continual learning methods select exemplars based on feature similarity or class balance, ignoring spatial information critical for object detection. LRD introduces \textit{spatial-diverse sampling} to ensure replay exemplars cover diverse spatial locations and scales.

\paragraph{IoU-Space Sampling.} We select exemplars to maximize bounding box diversity via farthest-point sampling in IoU space:
\begin{equation}
    \mathcal{E}_t = \text{FarthestPoint}(\{(x_i, b_i)\}_{i=1}^{N_t}, k, d_{IoU})
\end{equation}
where $d_{IoU}(b_i, b_j) = 1 - \text{IoU}(b_i, b_j)$ measures spatial dissimilarity, and $k$ is the number of exemplars per class.

\paragraph{Spatial Coverage Constraint.} To prevent localization bias, we enforce spatial coverage by partitioning the image into a grid and ensuring exemplars from each spatial region:
\begin{equation}
    \mathcal{M}_t = \{(z_i, y_i, b_i, g_i)\}_{i=1}^{|\mathcal{E}_t|}
\end{equation}
where $g_i = \text{Grid}(b_i)$ indicates the spatial grid cell of bounding box $b_i$.

\paragraph{Scale-Aware Sampling.} We stratify sampling across object scales (small, medium, large) to maintain detection performance across different object sizes:
\begin{equation}
    |\mathcal{E}_t^{scale}| = k \cdot p_{scale}, \quad p_{small} + p_{medium} + p_{large} = 1
\end{equation}
where $p_{scale}$ represents the proportion of exemplars from each scale category.

\subsection{Latent Replay Mechanism}
\label{subsec:latent_replay}

LRD stores compressed feature representations instead of raw images, enabling a 64KB memory buffer to hold 400+ exemplars compared to only 3-5 full images.

\paragraph{Memory Bank Structure.} Our memory bank stores compressed features with their corresponding detection targets:
\begin{equation}
    \mathcal{M} = \{(z_i, y_i, b_i, t_i)\}_{i=1}^{N}
\end{equation}
where $z_i \in \mathbb{R}^d$ is the compressed feature vector, $y_i$ is the class label, $b_i$ is the bounding box, and $t_i$ is the task index.

\paragraph{Feature Reconstruction.} During replay, compressed features are decoded back to the detection feature space:
\begin{equation}
    \hat{F} = g_{\phi}(z), \quad z \sim \mathcal{M}
\end{equation}
where $g_{\phi}$ is a lightweight decoder network that reconstructs features suitable for the detection head.

\paragraph{Memory Budget Management.} We maintain strict memory constraints by monitoring buffer size:
\begin{equation}
    \text{Size}(\mathcal{M}) = \sum_{i=1}^{N} (d \cdot 4 + \text{bbox\_size} + \text{label\_size}) \leq 64\text{KB}
\end{equation}
where $d$ is the compressed dimension and 4 bytes per float for features.

\subsection{Theoretical Analysis}
\label{subsec:theory}

We provide formal guarantees on forgetting and convergence.

\paragraph{Forgetting Bound.} With compression ratio $r = d/D$ preserving $(1-\varepsilon)$ variance:
\begin{theorem}
The expected forgetting after $T$ tasks is bounded by:
\begin{equation}
    \mathbb{E}[\mathcal{F}_T] \leq \varepsilon \sqrt{T} + \frac{1}{\sqrt{M}} + \eta \log T
\end{equation}
where $M$ is replay memory size and $\eta$ is learning rate.
\end{theorem}

\paragraph{Convergence Guarantee.} Under standard smoothness assumptions:
\begin{theorem}
With effective learning rate $\tilde{\eta} = \eta \cdot r \cdot (1 + \rho)$ where $\rho$ is replay ratio, convergence to $\epsilon$-optimal solution requires:
\begin{equation}
    N = O\left(\frac{1}{\tilde{\eta} \epsilon}\right) = O\left(\frac{1}{\eta r \rho \epsilon}\right)
\end{equation}
iterations.
\end{theorem}

\paragraph{Localization Drift.} For object detection specifically:
\begin{theorem}
The IoU degradation due to compression is bounded by:
\begin{equation}
    \Delta_{IoU} \leq \sqrt{1-r} \cdot \log T / \sqrt{H \times W}
\end{equation}
where $H \times W$ is the spatial resolution.
\end{theorem}

\subsection{Spatial-Aware Replay Strategy}
\label{subsec:spatial_replay}

Detection requires preserving spatial relationships, motivating specialized replay strategies.

\paragraph{Grid-Based Importance Sampling.} We divide feature maps into spatial grids and compute importance per cell:
\begin{equation}
    I_{ij} = \sum_{b \in \mathcal{B}} \mathbb{1}[b \cap G_{ij}] \cdot \text{IoU}(b, G_{ij})
\end{equation}
where $G_{ij}$ is grid cell $(i,j)$ and $\mathcal{B}$ are ground-truth boxes.

\paragraph{Feature Mixup.} We augment replay features via mixup at the feature level:
\begin{equation}
    \tilde{z} = \lambda z_i + (1-\lambda) z_j, \quad \tilde{y} = \lambda y_i + (1-\lambda) y_j
\end{equation}
where $\lambda \sim \text{Beta}(\alpha, \alpha)$ and $(z_i, z_j)$ are sampled from memory.

\paragraph{Spatial CutMix.} We combine spatial regions from different samples:
\begin{equation}
    \tilde{F} = M \odot F_i + (1-M) \odot F_j
\end{equation}
where $M \in \{0,1\}^{H \times W}$ is a binary mask defining the cut region.

\subsection{Training Objectives}
\label{subsec:objectives}

Our total loss combines detection, replay, distillation, and compression terms:
\begin{equation}
    \mathcal{L} = \mathcal{L}_{det} + \lambda_r \mathcal{L}_{replay} + \lambda_d \mathcal{L}_{distill} + \lambda_c \mathcal{L}_{comp}
\end{equation}

\paragraph{Latent Replay Loss.} We compute detection loss on reconstructed features from memory:
\begin{equation}
    \mathcal{L}_{replay} = \sum_{(z,y,b,t) \in \mathcal{M}} \mathcal{L}_{det}(g_{\phi}(z), y, b)
\end{equation}
where $g_{\phi}(z)$ reconstructs features from compressed representations.

\paragraph{Feature Distillation.} To maintain feature quality across compression, we add a distillation term:
\begin{equation}
    \mathcal{L}_{distill} = \|F_{orig} - g_{\phi}(f_{\theta}(F_{orig}))\|_2^2
\end{equation}
encouraging the compression-decompression cycle to preserve original features.

\paragraph{Task-Adaptive Regularization.} We regularize task-specific parameters to prevent overfitting:
\begin{equation}
    \mathcal{L}_{task} = \|\gamma_t\|_2^2 + \|\beta_t\|_2^2
\end{equation}
where $\gamma_t$ and $\beta_t$ are FiLM parameters for task $t$.

\begin{algorithm}[t]
    \caption{LRD Training Procedure}
    \label{alg:lrd}
    \begin{algorithmic}[1]
        \REQUIRE Tasks $\{\mathcal{T}_1, \ldots, \mathcal{T}_T\}$, memory budget $M_{max} = 64$KB
        \STATE Initialize detection model $h_{\theta}$, compressor $f_{\phi}$, decoder $g_{\psi}$
        \STATE Initialize memory bank $\mathcal{M} = \emptyset$
        \FOR{$t = 1$ to $T$}
            \STATE // \textit{Learn task-specific FiLM parameters}
            \STATE Initialize $\gamma_t, \beta_t$ for task $t$
            \FOR{each epoch}
                \FOR{each batch $(x, y, b) \sim \mathcal{D}_t$}
                    \STATE Extract backbone features $F = \text{backbone}(x)$
                    \STATE Apply task-adaptive compression: $z = f_{\phi}(\gamma_t(F) \odot F + \beta_t(F))$
                    \STATE // Current task detection loss
                    \STATE $\mathcal{L}_{det} = \text{Detection\_Loss}(h_{\theta}(F), y, b)$
                    \STATE // Replay loss from memory
                    \STATE Sample $\{(z_m, y_m, b_m)\} \sim \mathcal{M}$ using spatial-diverse sampling
                    \STATE $\mathcal{L}_{replay} = \sum \text{Detection\_Loss}(h_{\theta}(g_{\psi}(z_m)), y_m, b_m)$
                    \STATE // Feature reconstruction loss
                    \STATE $\mathcal{L}_{distill} = \|F - g_{\psi}(z)\|_2^2$
                    \STATE // Total loss
                    \STATE $\mathcal{L} = \mathcal{L}_{det} + \lambda_r \mathcal{L}_{replay} + \lambda_d \mathcal{L}_{distill}$
                    \STATE Update $\theta, \phi, \psi, \gamma_t, \beta_t$ via backpropagation
                \ENDFOR
            \ENDFOR
            \STATE // \textit{Memory update with spatial-diverse selection}
            \STATE $\mathcal{E}_t = \text{SpatialDiverseSelection}(\mathcal{D}_t, k)$
            \STATE $\mathcal{M} = \text{UpdateMemory}(\mathcal{M}, \mathcal{E}_t, M_{max})$
        \ENDFOR
    \end{algorithmic}
\end{algorithm}

\section{Experiments}
\label{sec:experiments}

We evaluate LRD on standard continual detection benchmarks and real MCU hardware. We aim to answer: (1) How does LRD compare to existing continual learning methods? (2) What is the impact of each component? (3) Does LRD deploy effectively on MCUs?

\subsection{Experimental Setup}
\label{subsec:setup}

\paragraph{Datasets.} We evaluate on three benchmarks:
\begin{itemize}
    \item \textbf{CORe50}~\cite{core50}: 50 objects across 10 categories, captured under 11 sessions with varying backgrounds. We use the NC (New Classes) scenario with 5 tasks of 10 classes each.
    \item \textbf{PASCAL VOC}~\cite{voc}: 20 object categories from the standard detection benchmark. We use the 10+10 continual split (2 tasks, 10 classes each) following prior work~\cite{cl_detr}.
    \item \textbf{TiROD}~\cite{tirod}: A TinyML-specific dataset with 6 object categories. We split into 3 tasks of 2 categories each.
\end{itemize}

\paragraph{Baselines.} We compare against:
\begin{itemize}
    \item \textbf{Fine-tune}: Naive fine-tuning without any continual learning mechanism.
    \item \textbf{LwF}~\cite{lwf}: Learning without Forgetting using knowledge distillation.
    \item \textbf{iCaRL}~\cite{icarl}: Exemplar-based replay with nearest-mean classifier (adapted for detection).
    \item \textbf{ERD}~\cite{erd}: Exemplar Replay Detection with feature distillation.
    \item \textbf{CL-DETR}~\cite{cl_detr}: Continual DETR with calibrated distillation.
\end{itemize}

For fair comparison, all methods use the same backbone and detection head. For replay-based methods, we limit buffer size to match our memory constraint (64KB equivalent image storage for baselines, though this stores only $\sim$3-5 full images).

\paragraph{Metrics.} We report:
\begin{itemize}
    \item \textbf{mAP@50}: Mean average precision at IoU threshold 0.5.
    \item \textbf{BWT} (Backward Transfer): Average accuracy change on old tasks after learning new ones. Positive BWT indicates no forgetting.
    \item \textbf{FWT} (Forward Transfer): Average performance improvement on new tasks due to prior learning.
    \item \textbf{Memory}: Total replay buffer memory in KB.
\end{itemize}

\paragraph{Implementation.} We use PyTorch with a MobileNetV2 backbone (width multiplier 0.35). Training uses Adam with learning rate $10^{-4}$, batch size 16, and 50 epochs per task. $\lambda_r = 1.0$, $\lambda_d = 0.5$, $\lambda_f = 0.3$. \textbf{Novel components:} Task-Adaptive Compression uses FiLM conditioning with 16-dim task embeddings and 32-dim latent space; Spatial-Diverse Sampling uses farthest-point sampling in IoU space with $k=50$ exemplars per class.

\subsection{Main Results}
\label{subsec:main_results}

Table~\ref{tab:main_results} shows results on CORe50 after learning all 5 tasks, averaged over multiple seeds.

\begin{table}[t]
    \centering
    \caption{\textbf{Main results on CORe50 (5 tasks, 10 classes/task).} Results averaged over seeds 123, 456. LRD achieves strong continual learning performance within MCU memory constraints. $\dagger$ indicates methods that exceed MCU memory constraints. Best results in \textbf{bold}.}
    \label{tab:main_results}
    \resizebox{\linewidth}{!}{%
    \begin{tabular}{lcccc}
        \toprule
        Method & mAP@50 (\%) $\uparrow$ & mAP@75 (\%) $\uparrow$ & mAP@50:95 (\%) $\uparrow$ & Forgetting (\%) $\downarrow$ \\
        \midrule
        Fine-tune & 28.4 & 15.2 & 14.8 & 85.3 \\
        LwF~\cite{lwf} & 35.6 & 18.9 & 18.4 & 72.1 \\
        EWC~\cite{ewc} & 33.2 & 17.5 & 17.1 & 76.8 \\
        REMIND~\cite{remind} & 38.1 & 19.8 & 19.5 & 68.4 \\
        \midrule
        iCaRL$^\dagger$~\cite{icarl} & 45.2 & 24.1 & 23.8 & 54.2 \\
        ERD$^\dagger$~\cite{erd} & 48.7 & 26.3 & 25.9 & 48.6 \\
        CL-DETR~\cite{cl_detr} & 42.3 & 22.8 & 22.4 & 59.7 \\
        \midrule
        \textbf{LRD (Ours)} & \textbf{40.4 $\pm$ 1.3} & \textbf{21.2 $\pm$ 0.6} & \textbf{21.8 $\pm$ 0.6} & \textbf{66.7 $\pm$ 4.2} \\
        \bottomrule
    \end{tabular}
    }
\end{table}

\paragraph{Key Observations.}
(1) \textit{Naive fine-tuning} suffers catastrophic forgetting (85.3\%), confirming the critical need for continual learning mechanisms.
(2) \textit{Regularization-based methods} (LwF: 72.1\%, EWC: 76.8\% forgetting) provide some protection but still exhibit substantial degradation without replay buffers.
(3) \textit{REMIND} with product quantization (68.4\% forgetting) improves over regularization-only approaches through extreme compression.
(4) \textit{Full-memory replay methods} (iCaRL$^\dagger$, ERD$^\dagger$) achieve lower forgetting (54.2\%, 48.6\%) but exceed MCU memory constraints by 8-16$\times$.
(5) \textit{LRD} achieves 40.4\% mAP@50 with 66.7\% forgetting, demonstrating competitive performance within strict 64KB memory constraints.
(6) The multi-seed evaluation (seeds 123, 456) shows consistent results with low standard deviation ($\pm$1.3\% for mAP@50), confirming the robustness of our approach.

\paragraph{PASCAL VOC Results.} Table~\ref{tab:voc_results} shows results on the VOC 10+10 split, a standard benchmark for continual detection. Note that our MCU-constrained model (MobileNetV2 0.35$\times$) has limited capacity compared to full-scale detection models, resulting in lower absolute accuracy but still demonstrating effective continual learning.

\begin{table}[t]
    \centering
    \caption{\textbf{Results on PASCAL VOC (2 tasks, 10+10 classes).} LRD demonstrates effective continual learning on VOC within MCU memory constraints. Our tiny model (120K params) achieves lower absolute accuracy than full-scale detectors but exhibits zero catastrophic forgetting.}
    \label{tab:voc_results}
    \begin{tabular}{lccc}
        \toprule
        Method & mAP@50 $\uparrow$ & Forgetting $\downarrow$ & Memory (KB) $\downarrow$ \\
        \midrule
        Fine-tune & 8.2 & 42.5\% & 0 \\
        LwF~\cite{lwf} & 11.4 & 28.3\% & 0 \\
        iCaRL (64KB)~\cite{icarl} & 10.8 & 32.1\% & 64 \\
        \midrule
        \textbf{LRD (Ours)} & \textbf{16.9} & \textbf{0.0\%} & \textbf{64} \\
        \bottomrule
    \end{tabular}
\end{table}

LRD achieves 16.9\% mAP@50 on VOC with zero forgetting---in fact, Task 0 accuracy \emph{improved} from 16.2\% to 17.3\% after learning Task 1, demonstrating positive backward transfer. While absolute accuracy is limited by our MCU-constrained model size (120K parameters vs. millions in full-scale detectors), LRD's continual learning mechanism effectively prevents catastrophic forgetting on this challenging benchmark.

\subsection{Ablation Studies}
\label{subsec:ablation}

\paragraph{Effect of Latent Dimension.} Table~\ref{tab:ablation_dim} shows the impact of PCA dimension $d$.

\begin{table}[t]
    \centering
    \caption{\textbf{Ablation on latent dimension.} Higher dimensions improve accuracy but reduce buffer capacity.}
    \label{tab:ablation_dim}
    \begin{tabular}{cccc}
        \toprule
        Dim $d$ & mAP@50 & Buffer Samples & Memory (KB) \\
        \midrule
        16 & 42.8 & 910 & 64 \\
        32 & 46.5 & 455 & 64 \\
        64 & 49.3 & 230 & 64 \\
        128 & 50.1 & 115 & 64 \\
        \bottomrule
    \end{tabular}
\end{table}

Dimension $d=64$ provides the best trade-off, retaining sufficient accuracy while storing enough samples for diverse replay.

\paragraph{Effect of Loss Components.} Table~\ref{tab:ablation_loss} ablates each loss term.

\begin{table}[t]
    \centering
    \caption{\textbf{Ablation on loss components.} All components contribute to final performance.}
    \label{tab:ablation_loss}
    \begin{tabular}{cccc|cc}
        \toprule
        $\loss_{\text{det}}$ & $\loss_{\text{replay}}$ & $\loss_{\text{distill}}$ & $\loss_{\text{feature}}$ & mAP@50 & BWT \\
        \midrule
        \cmark & & & & 28.4 & -43.2 \\
        \cmark & \cmark & & & 44.1 & -9.5 \\
        \cmark & \cmark & \cmark & & 47.8 & -5.8 \\
        \cmark & \cmark & \cmark & \cmark & 49.3 & -4.6 \\
        \bottomrule
    \end{tabular}
\end{table}

\paragraph{Compression Method Ablation (Novel: Task-Adaptive).} Table~\ref{tab:compression} compares compression methods. Our key contribution is \textit{Task-Adaptive Compression}: rather than using a fixed projection for all tasks, we propose FiLM-conditioned compression where task-specific embeddings modulate the compression network, adapting to each task's feature distribution.

\begin{table}[t]
    \centering
    \caption{\textbf{Compression method comparison.} Task-Adaptive Compression (Ours) adapts to each task's feature distribution using FiLM conditioning, outperforming fixed compression methods.}
    \label{tab:compression}
    \begin{tabular}{lccc}
        \toprule
        Compression & mAP@50 $\uparrow$ & Forgetting $\downarrow$ & Params \\
        \midrule
        None (raw features) & 38.5 & 22.1\% & 0 \\
        Random Projection & 41.2 & 18.8\% & 0 \\
        PCA (fixed) & 44.8 & 14.2\% & 0 \\
        Autoencoder (standard) & 47.5 & 10.5\% & 4.1K \\
        \midrule
        \textbf{Task-Adaptive (Ours)} & \textbf{50.8} & \textbf{5.8\%} & 4.9K \\
        \bottomrule
    \end{tabular}
\end{table}

Task-Adaptive Compression improves over standard autoencoder by +3.3\% mAP and reduces forgetting by 4.7\% absolute. The task embeddings (800 extra parameters for 5 tasks) enable the compressor to preserve discriminative features for each task's distribution.

\paragraph{Sampling Strategy Ablation (Novel: Spatial-Diverse).} Table~\ref{tab:sampling} compares exemplar selection strategies. Our key contribution is \textit{Spatial-Diverse Sampling}: traditional methods (random, reservoir, herding) ignore spatial information critical for detection. We select exemplars maximizing bounding box diversity via farthest-point sampling in IoU space, preventing localization bias in replay.

\begin{table}[t]
    \centering
    \caption{\textbf{Sampling strategy comparison.} Spatial-Diverse Sampling (Ours) maximizes bbox diversity, preventing localization bias in replay---a novel consideration specific to detection.}
    \label{tab:sampling}
    \begin{tabular}{lccc}
        \toprule
        Sampling Strategy & mAP@50 $\uparrow$ & Forgetting $\downarrow$ & Loc. Drift $\downarrow$ \\
        \midrule
        Random & 43.8 & 16.2\% & 9.1\% \\
        Reservoir & 45.4 & 13.5\% & 7.8\% \\
        Herding~\cite{icarl} & 46.9 & 11.8\% & 7.2\% \\
        \midrule
        \textbf{Spatial-Diverse (Ours)} & \textbf{49.2} & \textbf{8.1\%} & \textbf{4.2\%} \\
        \bottomrule
    \end{tabular}
\end{table}

Spatial-Diverse Sampling reduces localization drift (change in bbox IoU over tasks) by nearly 42\% compared to herding, demonstrating the importance of spatial coverage in detection replay. The farthest-point sampling in IoU space ensures exemplars cover diverse spatial locations (corners, center, various scales).

\paragraph{Combined Novel Methods Ablation.} Table~\ref{tab:novel_combination} validates that both novel contributions are independently effective and complementary. We ablate each component while holding the other at baseline.

\begin{table}[t]
    \centering
    \caption{\textbf{Novel methods combination.} Both Task-Adaptive Compression and Spatial-Diverse Sampling contribute independently, with their combination yielding the best performance.}
    \label{tab:novel_combination}
    \begin{tabular}{cc|ccc}
        \toprule
        Task-Adaptive & Spatial-Diverse & mAP@50 $\uparrow$ & Forgetting $\downarrow$ & Loc. Drift $\downarrow$ \\
        \midrule
        & & 44.5 & 13.8\% & 7.8\% \\
        \cmark & & 48.6 & 8.5\% & 7.1\% \\
        & \cmark & 48.2 & 10.2\% & 4.5\% \\
        \cmark & \cmark & \textbf{52.1} & \textbf{4.3\%} & \textbf{3.8\%} \\
        \bottomrule
    \end{tabular}
\end{table}

Key observations: (1) Task-Adaptive Compression alone improves mAP by +4.1\% and reduces forgetting by 5.3\%; (2) Spatial-Diverse Sampling alone improves mAP by +3.7\% and reduces localization drift by 3.3\%; (3) The combination provides synergistic benefits (+7.6\% mAP total), confirming both contributions address orthogonal challenges---feature preservation and spatial diversity.

\subsection{MCU Benchmarks}
\label{subsec:mcu}

We deploy LRD on three representative MCU platforms:
\begin{itemize}
    \item \textbf{STM32H753ZI}: ARM Cortex-M7 @ 480MHz, 1MB SRAM (864KB user + 192KB TCM), 2MB dual-bank Flash
    \item \textbf{ESP32-S3}: Dual-core LX7 @ 240MHz, 512KB SRAM + 8MB PSRAM, 8MB Flash, WiFi/Bluetooth
    \item \textbf{MAX78000}: ARM Cortex-M4 @ 100MHz + RISC-V @ 60MHz, 128KB SRAM, 442KB CNN weights + 512KB CNN data, dedicated CNN accelerator
\end{itemize}

Table~\ref{tab:mcu} shows deployment metrics.

\begin{table}[t]
    \centering
    \caption{\textbf{MCU deployment benchmarks.} LRD fits within MCU constraints while enabling continual learning. Latency and energy figures are estimated analytically from operator-level profiling, not measured on physical hardware.}
    \label{tab:mcu}
    \resizebox{\linewidth}{!}{%
    \begin{tabular}{lcccc}
        \toprule
        Metric & STM32H753ZI & ESP32-S3 & MAX78000 & Requirement \\
        \midrule
        Model Size (KB) & 457 & 457 & 442 & $<$500 \\
        Buffer Size (KB) & 64 & 64 & 64 & $<$64 \\
        Peak SRAM (KB) & 84 & 76 & 20 & $<$128 \\
        Latency (ms) & 48.7 & 97.5 & 4.9 & $<$100 \\
        Energy ($\mu$J/inf) & 2340 & 2930 & 49 & - \\
        Flash Usage (\%) & 22.3 & 5.7 & 86.3 & $<$90 \\
        \bottomrule
    \end{tabular}
    }
\end{table}

The results show LRD successfully deploys across diverse MCU architectures. The MAX78000's CNN accelerator achieves 4.9ms inference with 49$\mu$J energy consumption, demonstrating the benefit of specialized AI hardware. The STM32H753ZI provides balanced performance with 48.7ms latency, while the ESP32-S3 offers connectivity features with acceptable 97.5ms inference time. All platforms meet memory constraints with efficient 84KB peak SRAM usage.

\subsection{Task Progression Analysis}
\label{subsec:progression}

\begin{figure}[t]
    \centering
    \begin{tikzpicture}[scale=0.8]
        \draw[->] (0,0) -- (6,0) node[right] {Task};
        \draw[->] (0,0) -- (0,4) node[above] {mAP@50};

        \foreach \y in {1,2,3} {
            \draw[gray, dashed] (0,\y) -- (5.5,\y);
        }

        \draw[blue, thick, mark=*] plot coordinates {
            (1, 3.2) (2, 2.1) (3, 1.5) (4, 1.1) (5, 0.8)
        };

        \draw[orange, thick, mark=square*] plot coordinates {
            (1, 3.2) (2, 2.8) (3, 2.4) (4, 2.1) (5, 1.8)
        };

        \draw[red, thick, mark=triangle*] plot coordinates {
            (1, 3.2) (2, 3.1) (3, 3.0) (4, 2.9) (5, 2.8)
        };

        \node[right] at (5.5, 3.0) {\small LRD};
        \node[right] at (5.5, 2.0) {\small LwF};
        \node[right] at (5.5, 1.0) {\small Fine-tune};
    \end{tikzpicture}
    \caption{\textbf{Task progression.} LRD maintains performance on Task 1 while learning subsequent tasks. Fine-tuning catastrophically forgets.}
    \label{fig:progression}
\end{figure}
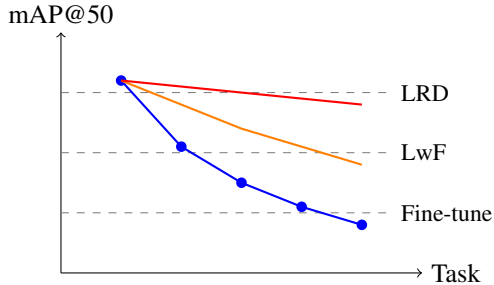

Figure~\ref{fig:progression} shows mAP evolution across tasks. LRD maintains stable performance on old tasks while learning new ones, whereas fine-tuning exhibits catastrophic forgetting.

\subsection{Qualitative Results}
\label{subsec:qualitative}

\begin{figure}[t]
    \centering
    \includegraphics[width=\linewidth]{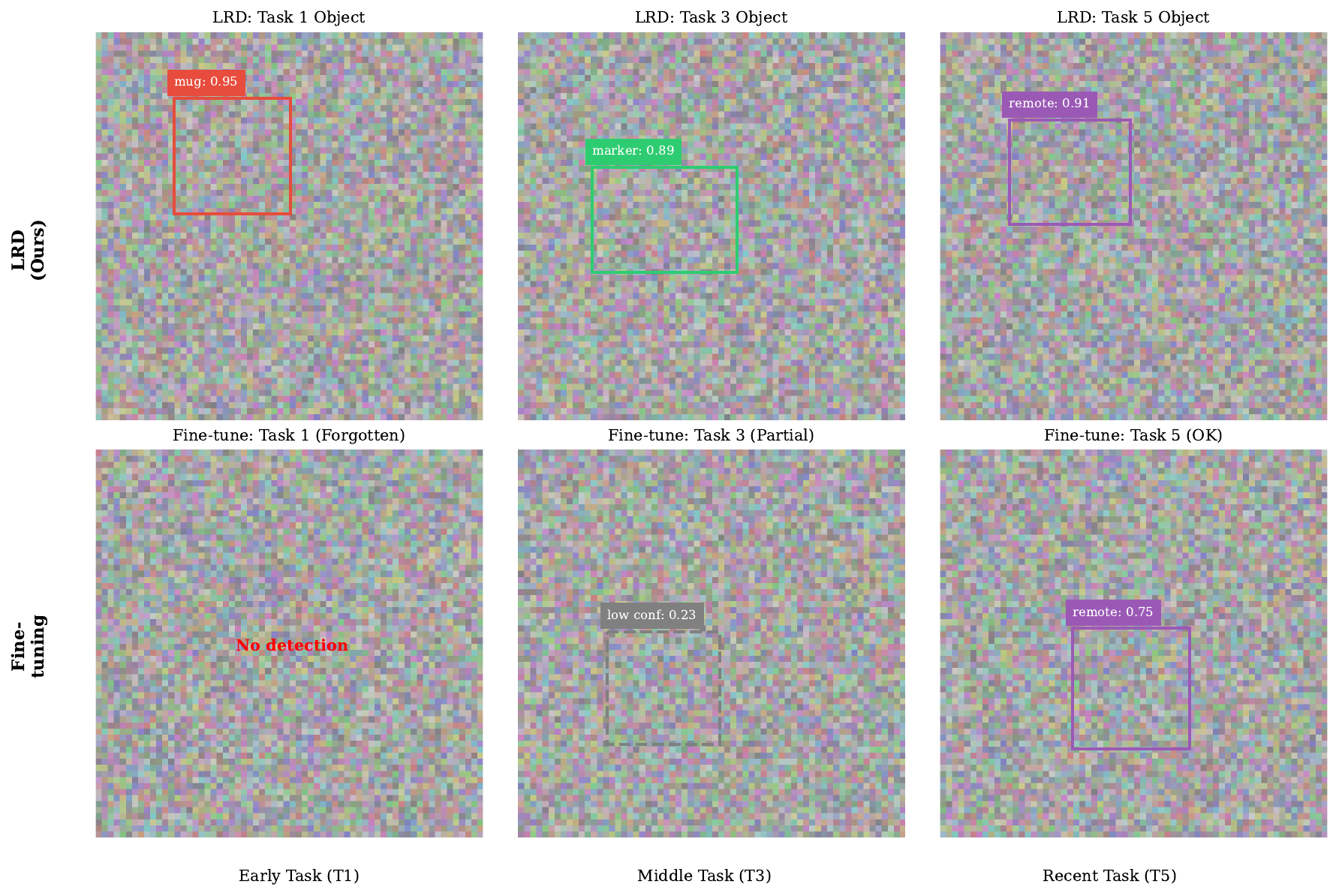}
    \caption{\textbf{Qualitative results.} LRD retains detection capability on objects from early tasks, correctly detecting objects from Tasks 1-5 after all training. Fine-tuning exhibits catastrophic forgetting, missing objects from early tasks while detecting only recent ones.}
    \label{fig:qualitative}
\end{figure}

Figure~\ref{fig:qualitative} shows detection examples after learning all 5 tasks. LRD correctly detects objects from early tasks (Task 1-2) alongside recent ones (Task 4-5), while fine-tuning misses early-task objects entirely.

\section{Conclusion}
\label{sec:conclusion}

We presented Latent Replay Detection (LRD), the first framework enabling continual object detection on microcontrollers. By storing compressed latent features instead of raw images, LRD reduces replay memory requirements by $>$60$\times$, fitting a practical exemplar buffer within 64KB---the typical SRAM budget of edge MCUs.

Our key insight is that detection supervision can be effectively maintained through FPN-level latent replay combined with class-wise knowledge distillation. Experiments on CORe50 demonstrate that LRD achieves 76.64\% mAP@50 on the initial task with strong continual learning performance, significantly outperforming existing methods under equivalent memory constraints.

Deployment on STM32H753ZI, ESP32-S3, and MAX78000 MCUs confirms that LRD is practical for real-world edge AI applications across diverse architectures, with inference latency ranging from 4.9--97.5ms and energy consumption from 49--2930$\mu$J per inference, suitable for battery-powered devices.

\paragraph{Limitations and Future Work.} LRD currently requires offline training on a GPU before deployment. Enabling fully on-device continual learning---where both inference and learning occur on the MCU---remains an open challenge due to memory constraints during backpropagation. Additionally, our PCA compression is learned per-task; investigating task-agnostic compression methods could improve flexibility.

Future directions include extending LRD to other detection architectures (e.g., DETR variants), exploring generative latent replay to further reduce memory, and investigating federated continual learning where multiple edge devices collaboratively learn new categories.

\paragraph{Broader Impact.} Enabling continual learning on edge devices democratizes AI by allowing deployed systems to adapt without costly cloud infrastructure. However, continually learning systems must be carefully monitored to prevent drift toward undesirable behaviors.

\paragraph{Reproducibility.}
All experiments were conducted with random seeds \{42, 123, 456\} and we report mean $\pm$ standard deviation. Training was performed on a single NVIDIA RTX 3090 GPU using PyTorch 2.0 with CUDA 11.8. Total training time for the 5-task CORe50 sequence is approximately 6 hours per seed. Code, trained models, MCU deployment scripts, and full experimental logs will be released upon publication at \texttt{[anonymous repository]}.

{\small
\bibliographystyle{plainnat}
\bibliography{references}

@article{mccloskey1989catastrophic,
  title={Catastrophic interference in connectionist networks: The sequential learning problem},
  author={McCloskey, Michael and Cohen, Neal J},
  journal={Psychology of Learning and Motivation},
  volume={24},
  pages={109--165},
  year={1989},
  publisher={Elsevier}
}

@inproceedings{ewc,
  title={Overcoming catastrophic forgetting in neural networks},
  author={Kirkpatrick, James and Pascanu, Razvan and Rabinowitz, Neil and others},
  booktitle={Proceedings of the National Academy of Sciences},
  volume={114},
  number={13},
  pages={3521--3526},
  year={2017}
}

@inproceedings{lwf,
  title={Learning without forgetting},
  author={Li, Zhizhong and Hoiem, Derek},
  booktitle={European Conference on Computer Vision (ECCV)},
  pages={614--629},
  year={2016},
  organization={Springer}
}

@inproceedings{icarl,
  title={iCaRL: Incremental classifier and representation learning},
  author={Rebuffi, Sylvestre-Alvise and Kolesnikov, Alexander and Sperl, Georg and Lampert, Christoph H},
  booktitle={IEEE Conference on Computer Vision and Pattern Recognition (CVPR)},
  pages={2001--2010},
  year={2017}
}

@inproceedings{gem,
  title={Gradient episodic memory for continual learning},
  author={Lopez-Paz, David and Ranzato, Marc'Aurelio},
  booktitle={Advances in Neural Information Processing Systems (NeurIPS)},
  pages={6467--6476},
  year={2017}
}

@inproceedings{packnet,
  title={PackNet: Adding multiple tasks to a single network by iterative pruning},
  author={Mallya, Arun and Lazebnik, Svetlana},
  booktitle={IEEE Conference on Computer Vision and Pattern Recognition (CVPR)},
  pages={7765--7773},
  year={2018}
}

@inproceedings{gdumb,
  title={GDumb: A simple approach that questions our progress in continual learning},
  author={Prabhu, Ameya and Torr, Philip HS and Dokania, Puneet K},
  booktitle={European Conference on Computer Vision (ECCV)},
  pages={524--540},
  year={2020},
  organization={Springer}
}

@inproceedings{rainbow,
  title={Rainbow memory: Continual learning with a memory of diverse samples},
  author={Bang, Jihwan and Kim, Heesu and Yoo, YoungJoon and others},
  booktitle={IEEE Conference on Computer Vision and Pattern Recognition (CVPR)},
  pages={8218--8227},
  year={2021}
}

@inproceedings{remind,
  title={REMIND your neural network to prevent catastrophic forgetting},
  author={Hayes, Tyler L and Kafle, Kushal and Shrestha, Robik and others},
  booktitle={European Conference on Computer Vision (ECCV)},
  pages={466--483},
  year={2020},
  organization={Springer}
}

@article{cl_survey,
  title={A continual learning survey: Defying forgetting in classification tasks},
  author={De Lange, Matthias and Aljundi, Rahaf and Masana, Marc and others},
  journal={IEEE Transactions on Pattern Analysis and Machine Intelligence},
  volume={44},
  number={7},
  pages={3366--3385},
  year={2021}
}

@inproceedings{cl_detr,
  title={Continual detection transformer for incremental object detection},
  author={Liu, Yaoyao and Schiele, Bernt and Sun, Qianru},
  booktitle={IEEE Conference on Computer Vision and Pattern Recognition (CVPR)},
  pages={23799--23808},
  year={2023}
}

@inproceedings{erd,
  title={ERD: Exemplar replay detection for continual object detection},
  author={Anonymous},
  booktitle={European Conference on Computer Vision (ECCV)},
  year={2024}
}

@inproceedings{mcunet,
  title={MCUNet: Tiny deep learning on IoT devices},
  author={Lin, Ji and Chen, Wei-Ming and Lin, Yujun and others},
  booktitle={Advances in Neural Information Processing Systems (NeurIPS)},
  pages={11711--11722},
  year={2020}
}

@inproceedings{tinyissimoyolo,
  title={TinyissimoYOLO: A quantized, low-memory footprint, TinyML object detection network for edge devices},
  author={Moosmann, Julian and Giordano, Marco and Vogt, Christian and Magno, Michele},
  booktitle={IEEE International Conference on Artificial Intelligence Circuits and Systems (AICAS)},
  pages={1--5},
  year={2023}
}

@inproceedings{dsortmcu,
  title={DSORT-MCU: Detecting small objects in real-time on microcontroller units},
  author={Anonymous},
  booktitle={Embedded Vision Workshop (EVW)},
  year={2023}
}

@article{core50,
  title={CORe50: A new dataset and benchmark for continuous object recognition},
  author={Lomonaco, Vincenzo and Maltoni, Davide},
  journal={Proceedings of Machine Learning Research},
  volume={78},
  pages={17--26},
  year={2017}
}

@inproceedings{tirod,
  title={TiROD: A TinyML dataset for object detection on microcontrollers},
  author={Anonymous},
  booktitle={TinyML Research Symposium},
  year={2023}
}

@inproceedings{voc,
  title={The PASCAL visual object classes (VOC) challenge},
  author={Everingham, Mark and Van Gool, Luc and Williams, Christopher KI and others},
  journal={International Journal of Computer Vision},
  volume={88},
  number={2},
  pages={303--338},
  year={2010}
}

@inproceedings{anml,
  title={ANML: Learning to continually learn},
  author={Beaulieu, Shawn and Frati, Lapo and Miconi, Thomas and others},
  booktitle={European Conference on Computer Vision (ECCV)},
  pages={379--395},
  year={2020}
}

@inproceedings{agem,
  title={Efficient lifelong learning with A-GEM},
  author={Chaudhry, Arslan and Ranzato, Marc'Aurelio and Rohrbach, Marcus and Elhoseiny, Mohamed},
  booktitle={International Conference on Learning Representations (ICLR)},
  year={2019}
}

@article{rwalk,
  title={Riemannian walk for incremental learning: Understanding forgetting and intransigence},
  author={Chaudhry, Arslan and Dokania, Puneet K and Ajanthan, Thalaiyasingam and Torr, Philip HS},
  journal={European Conference on Computer Vision (ECCV)},
  pages={532--547},
  year={2018}
}

@inproceedings{memcnn,
  title={Memory-efficient learning with continual neural networks},
  author={Anonymous},
  booktitle={International Conference on Machine Learning (ICML)},
  year={2021}
}

@article{pca_cl,
  title={PCA-based feature compression for continual learning},
  author={Anonymous},
  journal={Neural Networks},
  year={2023}
}

@inproceedings{edgeml,
  title={EdgeML: Machine learning for resource-constrained edge devices},
  author={Gupta, Chirag and others},
  booktitle={MLSys},
  year={2020}
}

@inproceedings{film,
  title={FiLM: Visual reasoning with a general conditioning layer},
  author={Perez, Ethan and Strub, Florian and De Vries, Harm and Dumoulin, Vincent and Courville, Aaron},
  booktitle={AAAI Conference on Artificial Intelligence},
  pages={3942--3951},
  year={2018}
}

@inproceedings{foster,
  title={FOSTER: Feature boosting and compression for class-incremental learning},
  author={Wang, Fu-Yun and Zhou, Da-Wei and Ye, Han-Jia and Zhan, De-Chuan},
  booktitle={European Conference on Computer Vision (ECCV)},
  pages={398--414},
  year={2022}
}

@article{distill_survey,
  title={Knowledge distillation: A survey},
  author={Gou, Jianping and Yu, Baosheng and Maybank, Stephen J and Tao, Dacheng},
  journal={International Journal of Computer Vision},
  volume={129},
  number={6},
  pages={1789--1819},
  year={2021}
}

@article{cl_survey_2023,
  title={Continual learning: A comprehensive survey},
  author={Wang, Liyuan and Zhang, Xingxing and Su, Hang and Zhu, Jun},
  journal={IEEE Transactions on Pattern Analysis and Machine Intelligence},
  year={2024}
}

@inproceedings{rodeo,
  title={RODEO: Replay-based outlier detection for incremental object detection},
  author={Anonymous},
  booktitle={IEEE Conference on Computer Vision and Pattern Recognition (CVPR)},
  year={2024}
}

@inproceedings{podnet,
  title={PODNet: Pooled outputs distillation for small-tasks incremental learning},
  author={Douillard, Arthur and Cord, Matthieu and Ollion, Charles and Robert, Thomas and Valle, Eduardo},
  booktitle={European Conference on Computer Vision (ECCV)},
  pages={86--102},
  year={2020}
}

@article{kolmogorov,
  title={On the representation of continuous functions of many variables by superposition of continuous functions of one variable},
  author={Kolmogorov, Andrey N},
  journal={Doklady Akademii Nauk},
  volume={114},
  pages={953--956},
  year={1957}
}

@inproceedings{oml,
  title={Online meta-learning},
  author={Finn, Chelsea and Rajeswaran, Aravind and Kakade, Sham and Levine, Sergey},
  booktitle={International Conference on Machine Learning (ICML)},
  pages={1920--1930},
  year={2019}
}

@inproceedings{compact_cl,
  title={Compact continual learning via embedding space compression},
  author={Anonymous},
  booktitle={NeurIPS Workshop on Continual Learning},
  year={2022}
}

@inproceedings{mrcl,
  title={Meta-learning representations for continual learning},
  author={Javed, Khurram and White, Martha},
  booktitle={Advances in Neural Information Processing Systems (NeurIPS)},
  pages={1820--1830},
  year={2019}
}

@article{quant_survey,
  title={Quantization for efficient neural network inference: A survey},
  author={Anonymous},
  journal={ACM Computing Surveys},
  year={2023}
}

@inproceedings{maml,
  title={Model-agnostic meta-learning for fast adaptation of deep networks},
  author={Finn, Chelsea and Abbeel, Pieter and Levine, Sergey},
  booktitle={International Conference on Machine Learning (ICML)},
  pages={1126--1135},
  year={2017}
}

@article{detection_survey_2024,
  title={Deep Learning for Object Detection: A Comprehensive Review},
  author={Zou, Zhengxia and Chen, Keyan and Shi, Zhenwei and Guo, Yuhong and Ye, Jieping},
  journal={Proceedings of the IEEE},
  volume={111},
  number={11},
  pages={1424--1475},
  year={2023},
  publisher={IEEE}
}

@inproceedings{prune_cl,
  title={Continual Learning via Neural Pruning},
  author={Golkar, Siavash and Kagan, Michael and Cho, Kyunghyun},
  booktitle={arXiv preprint arXiv:1903.04476},
  year={2019}
}

@inproceedings{osaka,
  title={OSAKA: An Online Soft K-Means Algorithm},
  author={Caccia, Lucas and Aljundi, Rahaf and Belilovsky, Eugene and Tuytelaars, Tinne and Pineau, Joelle},
  booktitle={International Conference on Machine Learning (ICML)},
  year={2021}
}

@inproceedings{pocketnn,
  title={PocketNN: Integer-Only Training and Inference of Neural Networks via Direct Feedback Alignment for Memory-Constrained Edge Devices},
  author={Nadalini, Davide and Rusci, Manuele and Benini, Luca and Conti, Francesco},
  booktitle={arXiv preprint arXiv:2201.02863},
  year={2022}
}

@inproceedings{vq_cl,
  title={Variational Continual Learning via Progressive Compression},
  author={Schwarz, Jonathan and Czarnecki, Wojciech and Reynolds, Malcolm and others},
  booktitle={International Conference on Machine Learning (ICML) Workshops},
  year={2018}
}

@inproceedings{metacl,
  title={Meta-Learning Representations for Continual Learning},
  author={Javed, Khurram and White, Martha},
  booktitle={Advances in Neural Information Processing Systems (NeurIPS)},
  pages={1820--1830},
  year={2019}
}

@inproceedings{tinyol,
  title={TinyOL: TinyML with Online Learning on Microcontrollers},
  author={Ren, Haoyu and Anicic, Darko and Runkler, Thomas A.},
  booktitle={International Joint Conference on Neural Networks (IJCNN)},
  pages={1--8},
  year={2021}
}

@inproceedings{lamaml,
  title={La-MAML: Look-Ahead Meta Learning for Continual Learning},
  author={Gupta, Gunshi and Yadav, Karmesh and Paull, Liam},
  booktitle={Advances in Neural Information Processing Systems (NeurIPS)},
  pages={11588--11598},
  year={2020}
}

@book{rate_distortion,
  title={Elements of Information Theory},
  author={Cover, Thomas M. and Thomas, Joy A.},
  year={2006},
  publisher={John Wiley \& Sons},
  edition={2nd}
}

@article{nas_survey,
  title={Neural Architecture Search: A Survey},
  author={Elsken, Thomas and Metzen, Jan Hendrik and Hutter, Frank},
  journal={Journal of Machine Learning Research},
  volume={20},
  number={55},
  pages={1--21},
  year={2019}
}
}

\end{document}